\documentclass[11pt,a4paper]{article}
\usepackage[hyperref]{emnlp2020}
\usepackage{times}
\usepackage{latexsym}

\usepackage{microtype}

\usepackage{graphicx}

\usepackage{algorithm}  
\usepackage{algorithmicx} 
\usepackage{algpseudocode}

\usepackage{multirow}
\usepackage{booktabs}
\usepackage[normalem]{ulem}
\useunder{\uline}{\ul}{}
\usepackage{colortbl} 
\usepackage{arydshln} 

\usepackage{amsmath}
\usepackage{amssymb}
\usepackage{verbatim}
\usepackage{siunitx}
\usepackage{bm}
\let\vec\boldsymbol

\usepackage{color}
\usepackage{nicefrac}
\usepackage{appendix}

\newcommand{\myparagraph}[1]{\paragraph{#1}}

\aclfinalcopy 

\title{Regularizing Dialogue Generation by Imitating Implicit Scenarios}

\author{Shaoxiong Feng$^{\text{1}}$\enskip Xuancheng Ren$^{\text{2}}$\enskip Hongshen Chen$^{\text{3}}$\enskip Bin Sun$^{\text{1}}$\enskip Kan Li$^{\text{1}}$\enskip
Xu Sun$^{\text{2}}$\\
$^{\text{1}}$Beijing Institute of Technology\\ 
$^{\text{2}}$MOE Key Laboratory of Computational Linguistics, School of EECS, Peking University\\ $^{\text{3}}$JD.com\\
{\small \tt \{shaoxiongfeng, binsun, likan\}@bit.edu.cn \{renxc, xusun\}@pku.edu.cn} \\
{\small \tt ac@chenhongshen.com}}

\date{}

\begin{document}
\maketitle

\begin{abstract}
Human dialogues are scenario-based and appropriate responses generally relate to the latent context knowledge entailed by the specific scenario. To enable responses that are more meaningful and context-specific, we propose to improve generative dialogue systems from the scenario perspective, where both dialogue history and \textit{future conversation} are taken into account to implicitly reconstruct the scenario knowledge. More importantly, the conversation scenarios are further internalized using imitation learning framework, where the conventional dialogue model that has no access to future conversations is effectively regularized by transferring the scenario knowledge contained in hierarchical supervising signals from the scenario-based dialogue model, so that the future conversation is not required in actual inference. Extensive evaluations show that our approach significantly outperforms state-of-the-art baselines on diversity and relevance, and expresses scenario-specific knowledge.
\end{abstract}

\section{Introduction} 
Neural dialogue generation has drawn increasing attention due to its vast commercial values and practical demands. Typically, given the dialogue history, neural dialogue models, such as plain Seq2Seq model \citep{DBLP:conf/nips/SutskeverVL14} and Transformer \citep{DBLP:conf/nips/VaswaniSPUJGKP17}, learn to predict responses via maximum likelihood estimation \citep{DBLP:journals/corr/VinyalsL15,DBLP:conf/acl/ShangLL15}.

\begin{figure}[t]
    \centering
    \includegraphics[width=1.0\linewidth]{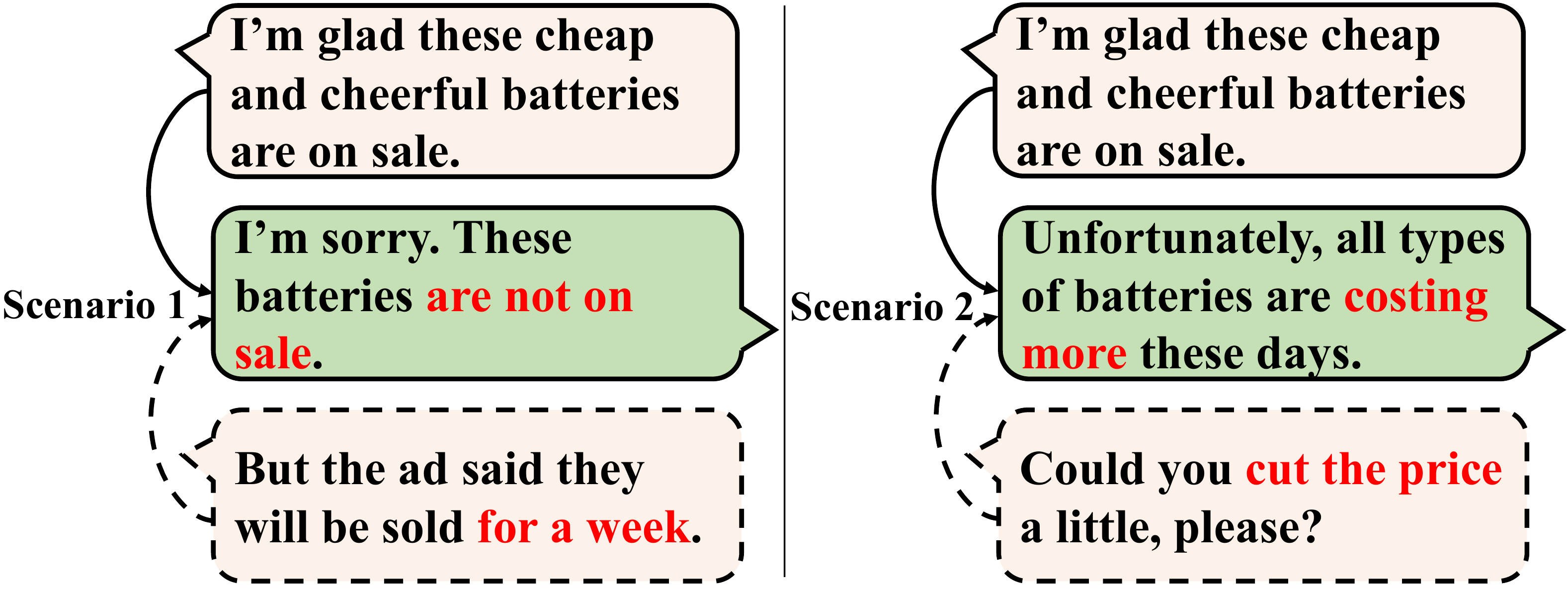}
    \caption{Examples of one dialogue history and its different responses followed by related future conversations. Different responses imply various conversation scenarios, which can be inferred by different relevant future conversations.} 
    \label{tab:intro_figure}
\end{figure}

Different from other sequence generation tasks, such as machine translation and paraphrase generation, the dialogue generation task can be regarded as a loose-coupling task, which has much freedom in the semantic and the linguistic aspects of the generated responses. However, it is often hard for the existing models to handle such freedom, compared to the fact that humans have no problem in giving specific yet varied responses even for open-ended dialogue history \citep{DBLP:conf/emnlp/ShenSLK18,DBLP:conf/acl/CsakyPR19}. One important reason is that we can extend the given dialogue with many possible scenarios of enriched, imaginative background information from our experience and world knowledge, to which existing systems have no access. 

It is beneficial for the dialogue systems to build upon such scenarios to facilitate dialogue generation. However, manually annotating the scenario contexts is intractable in terms of both difficulty and quantity. In turn, we find that such scenarios are naturally  contained in existing multi-turn dialogue corpora, where the entire dialogue of both dialogue history and future conversation with respect to the current utterance implicitly represents a specific dialogue scenario. An example is given in Figure~\ref{tab:intro_figure}. For Scenario 1,  ``for a week'' in the future conversation suggests the response is related to time. For Scenario 2, ``cut the price'' indicates that the response contains price information. 

Therefore, we reconstruct the dialogue task that only relies on dialogue history into a scenario-based response generation task. In order to enrich the conversation scenario, we employ future conversations together with dialogue histories to learn implicit conversation scenarios, which provide more semantic constraints to guide the response generation. We further propose a novel model to handle this new type of training data consisting of \textit{\{implicit scenario, response\}} pairs. 

It should be noted that the scenario-based dialogue model relies on future conversations that are inaccessible in inference. Rather than simply searching the training corpora for possible scenarios, we propose an imitation learning framework to drive the conventional dialogue model to absorb the corresponding scenario knowledge from the scenario-based dialogue model. Specifically, the scenario-based dialogue model serves as a teacher, and the conventional dialogue model that relies solely on dialogue history serves as a student that mimics the outputs of the teacher. Under the regularization of scenario knowledge, the student is effectively guided towards a wider local minimum that represents better generalization performance \citep{DBLP:conf/iclr/ChaudhariCSLBBC17,DBLP:conf/iclr/KeskarMNST17}. To facilitate knowledge transfer, the student mimics the teacher on every layer instead of just the top layer, which alleviates the delayed supervised signal problem using hierarchical semantic information in the teacher \citep{DBLP:journals/corr/abs-1909-06708}. Besides containing the information of future conversations, the distilled knowledge \citep{DBLP:journals/corr/HintonVD15} is also a less noisy and more ``deterministic'' supervised signal in comparison to real-world responses \citep{DBLP:conf/emnlp/LeeMC18,DBLP:conf/aaai/GuoTHQXL19}, which provides the student with smoother sequence trajectories that are easier to fit. 

We highlight our contributions as follows:
\begin{itemize}
\item We introduce future conversations together with dialogue histories to learn implicit conversation scenarios, which provide more semantic constraints to drive the responses to be meaningful and relevant to the real-world scenario-specific knowledge. 
\item We propose an imitation learning framework that bridges the gap between training and inference in the accessibility of future conversations. We also demonstrate why imitation learning works and further how to enhance the imitation learning. 
\item Our model achieves better results than state-of-the-art baselines on four datasets. Extensive analysis demonstrates the effectiveness and the scalability of the implicit conversation scenarios and the proposed imitation learning framework.
\end{itemize}

\section{Proposed Approach}
\begin{figure}[t]
    \centering
    \includegraphics[width=1.0\linewidth]{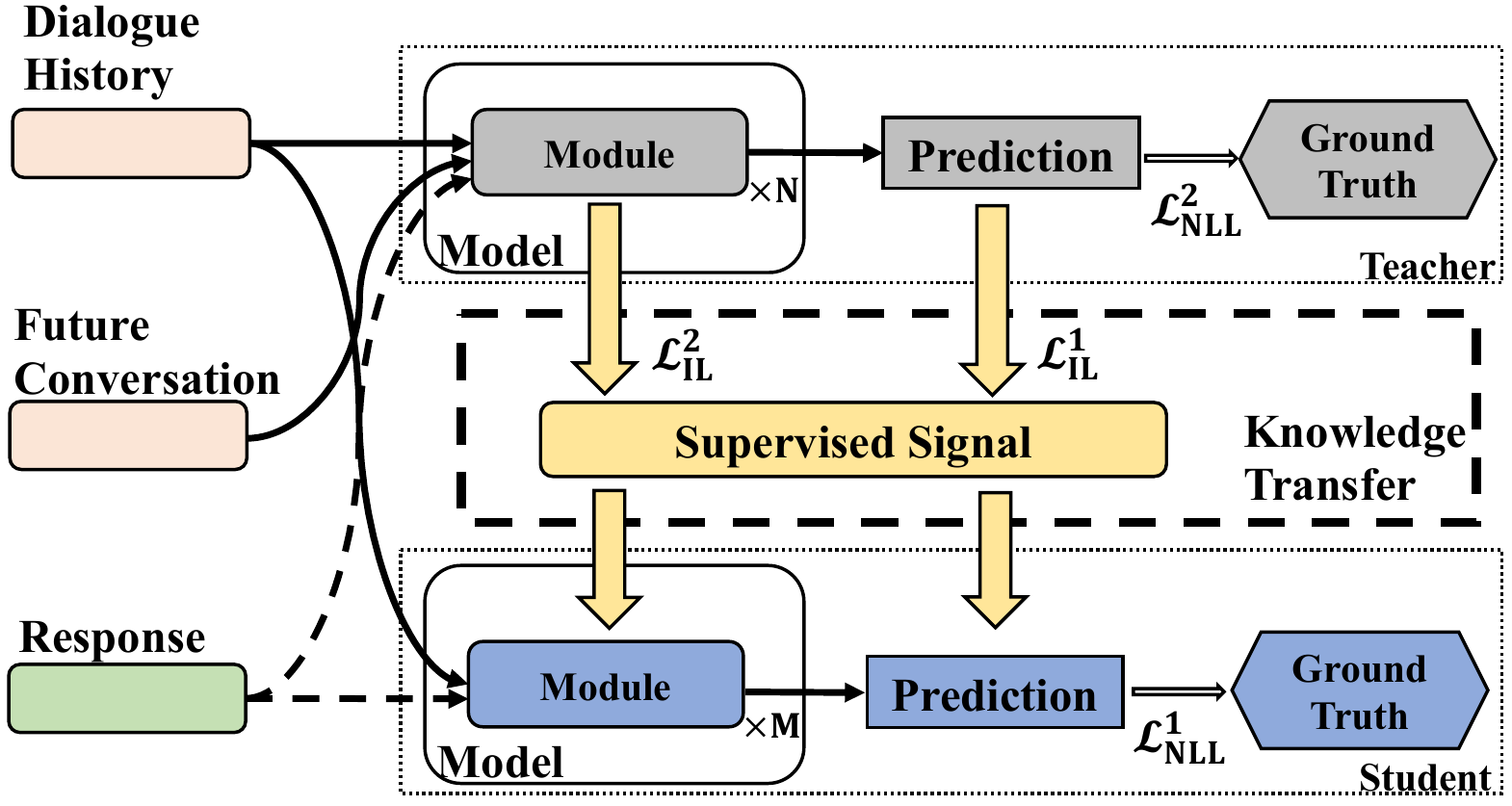}
    \caption{Illustration of the imitation learning framework transferring scenario knowledge from the teacher to the student. Top (Teacher): the scenario-based dialogue model. Bottom (Student): the conventional dialogue model.}
    \label{tab:architecture}
\end{figure}

In this section, we first introduce the scenario-based dialogue model, then describe the imitation learning framework shown in Figure \ref{tab:architecture}, and finally present the training objective.

\subsection{Scenario-Based Dialogue Model}
The conventional dialogue model takes a sequence of dialogue history $X = \{x_{1}, \ldots, x_{T}\}$ as input, and generates a response $Y = \{y_{1}, \ldots, y_{T^{'}}\}$ word by word, where $T$ and $T^{'}$ represent the length of source side and target side respectively. The maximum likelihood estimation is usually used to train the model, which can also be expressed as minimizing the negative log-likelihood:
\begin{equation}
\small
\begin{aligned}
\mathcal{L}^{1}_{\mathrm{NLL}}(\theta_{1}) = & 
 -\sum_{i=1}^{T^{'}}\sum_{k=1}^{|\mathcal{V}|}\  \mathbb{I}\{y_{i}=k\} \cdot \\
& \log p ( y_{i}=k |y_{<i}, X_{h} ; \theta_{1} ),
\end{aligned}
\label{equation:NLL 1}
\end{equation}
where $|\mathcal{V}|$ is the size of vocabulary, $\theta_{1}$ is a set of parameters, and $X_{h}$ represents the input sequence that is from dialogue history.

However, dialogue task allows responses to continue the dialogue topic from many aspects or even introduce a new topic depending on various conversation scenarios or semantic constraints, which dramatically increases the difficulty of prediction without any specific scenario information besides the hints from the given dialogue history. Moreover, labeling the scarce scenario information is labor-consuming and impractical. Instead, we resort to easy-to-access but underutilized future conversations that exist in all multi-turn dialogue corpora. By combining the dialogue history and its corresponding future conversation, we introduce the implicit conversation scenario into existing dialogue models to provide more semantic constraints and reduce the difficulty of prediction.

Concretely, we enforce the model to use implicit conversation scenarios to generate responses from two aspects. Different from previous dialogue models only based on the dialogue history $X_{h}$ to predict the response $Y$, the future conversation $X_{f}$ is also considered as part of input so that the model can look ahead and predict more purposefully. Intuitively, our training pair $\{(X_{h}, X_{f}), Y\}$ induces the model to imitate humans to produce the scenario-specific response. 

We also redesign the sequence generation architecture to handle the proposed training pair. The attention module in each layer calculates the weight of the contextualized token representations from the encoder based on the information that has been generated in the decoder, and then returns the context $c_{h}$. In order to consider the future conversation $X_{f}$, we apply another encoder to produce the contextualized token representations of $X_{f}$, which will be further extracted as the context $c_{f}$ by the attention module. The new encoder shares the parameters with the original encoder. Meanwhile, the output of the attention module is the concatenation of the past context $c_{h}$ and the future context $c_{f}$. Finally, the training criterion is formulated as the following negative log-likelihood: 
\begin{equation}
\small
\begin{aligned}
\mathcal{L}^{2}_{\mathrm{NLL}}(\theta_{2}) = &
 -\sum_{i=1}^{T^{'}}\sum_{k=1}^{|\mathcal{V}|}\ 
\mathbb{I}\{y_{i}=k\} \cdot \\
& \log p ( y_{i}=k |y_{<i}, X_{h}, X_{f} ; \theta_{2} ),
\end{aligned}
\label{equation:NLL 2}
\end{equation}
where $\theta_{2}$ is a set of parameters to minimize the NLL loss for the scenario-based dialogue model.

\subsection{Imitation Learning}
In inference, future conversations are inaccessible, which means implicit conversation scenarios cannot be constructed. Thus, the performance improvement from the scenario-based dialogue model cannot facilitate the generation of high-quality responses in practice. In order to bridge this gap between training and inference, we propose an imitation learning framework, in which we regard the scenario-based dialogue model as a teacher and the conventional dialogue model as a student. Through step-by-step imitation, including fine-grained prediction imitation and intermediate representation imitation, scenario knowledge distilled from the teacher regularizes the student to reach a robust local minimum and obtain significant generalization performance in inference.

\subsubsection{Fine-Grained Prediction Imitation}
Compared with the ground-truth labels, the soft predictions (i.e., the probability distribution from the output layer) contain more fine-grained and valuable information, such as the similarity of labels and potential future conversations. Moreover, the soft predictions provide less noisy and more ``deterministic'' targets that are easy to mimic. To transfer knowledge from the teacher, instead of taking the one-hot representation of $Y$ as the target, we minimize the cross-entropy of the predicted probability distribution between the teacher and the student:
\begin{equation}
\small
\begin{aligned}
\mathcal{L}^{1}_{\mathrm{IL}} (\theta_{1},\theta_{2}) = -\sum_{i=1}^{T^{'}}\sum_{k=1}^{|\mathcal{V}|}\ & p ( y_{i}=k |y_{<i}, X_{h}, X_{f} ; \theta_{2} ) \cdot \\
& \log p ( y_{i}=k |y_{<i}, X_{h} ; \theta_{1} )
\end{aligned}
\label{equation:IL 1}
\end{equation}

\subsubsection{Intermediate Representation Imitation} 
Only transferring knowledge from the output layer has a limited effect on the student to use implicit conversation scenarios. When the student network is very deep, the supervised signals from the output layer hardly conduct an effective update and regularization on the parameters of intermediate layers, which will make the imitation learning framework quickly reach saturation \citep{DBLP:journals/corr/RomeroBKCGB14,DBLP:journals/corr/abs-1908-09355}. This problem prevents the student from scaling to deeper models to further improve the model performance. 

To tackle this problem, we extend the range of imitation learning from the soft predictions in the output layer to the output $h$ of intermediate layers to guide the imitation process. Specifically, we penalize the discrepancy of hidden states in intermediate layers between the teacher and the student:
\begin{equation}
\small
\begin{aligned}
\mathcal{L}^{2}_{\mathrm{IL}} (\theta_{1},\theta_{2}) = 
 \sum_{i=1}^{T^{'}} \sum_{l=1}^{|\mathcal{O}|} f\big(  & h^{t}_{il}(X_{h},X_{f} ; \theta_{2}), \\
 & h^{s}_{il}(X_{h};\theta_{1}) \big),
\end{aligned}
\label{equation:IL 2}
\end{equation}
where $|\mathcal{O}|$ is the number of intermediate layers, $h^{t}_{il}$ and $h^{s}_{il}$ are the outputs of intermediate layers in the teacher and the student respectively, and $f(\cdot)$ is the measurement function. 
\begin{equation}
\small
f \left( \cdot \right) = 
\begin{cases}
\phi(h^{s}_{il},h^{t}_{il}), 
& \text{ if } \phi(h^{s}_{il},h^{t}_{il}) \geq \alpha ; \\
0,
& \text{ else } .
\end{cases}
\end{equation}
where $\phi(\cdot)$ is the mean-squared-error (MSE) loss. Because we observe that directly applying the MSE loss as an additional loss hurts the stability of the imitation learning process, we set a scalar threshold $\alpha$ to loose this constraint.

\subsection{Training} 
Combining the NLL loss in Equation (\ref{equation:NLL 1}) with the IL losses in Equation (\ref{equation:IL 1}) and Equation (\ref{equation:IL 2}), the final objective function of the student is formulated as:
\begin{equation}
\small
\mathcal{L} = \mathcal{L}^{1}_{NLL} + \lambda_{1} (\mathcal{L}^{1}_{\mathrm{IL}} + \mathcal{L}^{2}_{\mathrm{IL}}),
\label{equation:final obejctive}
\end{equation}
where $\lambda_{1}$ is a hyper-parameter that balances the importance of the NLL loss and the IL losses.

Because the scenario knowledge is only transferred from the teacher by hierarchical supervised signals, our imitation framework has the following three advantages: (1) Compared with the fine-tuning style of knowledge transfer \citep{DBLP:conf/nips/DaiL15,DBLP:conf/acl/RuderH18}, the proposed imitation framework does not affect the teacher, i.e., the knowledge learned from the teacher will not be forgotten. (2) The proposed method is model agnostic. Thus, the imitation object can be extended from one teacher to multiple teachers, such as incorporating a language model besides the scenario-based dialogue model. (3) The imitation process does not change the current objective function, which means the previous work of modifying objective function can serve as a complementary to improve the model performance further.

\section{Experiment}

\subsection{Datasets}
\myparagraph{DailyDialog} It is provided by \citet{DBLP:conf/ijcnlp/LiSSLCN17}, which contains various dialogue topics about daily life. We randomly select 27K, 2.5K, and 1.5K pairs for training, validation, and testing. 

\myparagraph{PersonaChat} It is gathered by assigning two Amazon Turkers with their personas to chat with each other \citep{DBLP:conf/acl/KielaWZDUS18}. We only use the conversation section and split it to 67K, 8.5K, and 8K pairs for training, validation, and testing. 

\myparagraph{OpenSubtitles} It is collected from movie subtitles and consists of more than 60M scripted lines \citep{DBLP:conf/lrec/LisonT16}. We randomly extract 1500K, 50K, and 25K pairs for training, validation, and testing. 

For all datasets, every seven consecutive dialogue turns form a training example, in which the first three turns, the middle turn, and the last three turns are taken as \textit{dialogue history}, \textit{response}, and \textit{future conversation}, respectively. 

We also conducted the experiment on a multi-domain goal-oriented dataset called \textbf{MultiWOZ}, which is simplified by us as a general dialogue generation task. The detailed description of MultiWOZ and data pre-processing is provided in Appendix~\ref{ap:datasets}.

\subsection{Baselines}
We re-implemented two classes of six baselines for comparison. The detailed settings of baselines are provided in Appendix~\ref{ap:baselines}.

\myparagraph{LSTM-Based} 
One class is based on LSTM, including \textbf{Seq2Seq+Att}, which contains a vanilla Seq2Seq model \citep{DBLP:conf/nips/SutskeverVL14} with attention mechanism \citep{DBLP:journals/corr/BahdanauCB14}, \textbf{VHRED+BOW} \citep{DBLP:conf/aaai/SerbanSLCPCB17}, which introduces a continuous latent variable attached to the response information into HRED \citep{DBLP:conf/aaai/SerbanSBCP16} and applies BOW loss \citep{DBLP:conf/acl/ZhaoZE17} as a complementary with KL annealing, and \textbf{NEXUS} \citep{DBLP:conf/emnlp/ShenSLK18}, which further uses the future conversation to incorporate more scenario information into the latent variable. 

\myparagraph{Transformer-Based} 
The other class is based on \textbf{Transformer} \citep{DBLP:conf/nips/VaswaniSPUJGKP17}, including itself, \textbf{ReCoSa} \citep{DBLP:conf/acl/ZhangLPGC19}, and \textbf{CHMAM} \citep{DBLP:conf/ijcai/TaoGSWZY18}, which consists of Multi-Head Attention Mechanism (MHAM) and an attention weight regularizer. Both ReCoSa and CHMAM aim to extract more relevant and diverse scenario information from dialogue history.

\subsection{Experiment Settings}
Based on the performance including the loss and metrics on the validation dataset, we trained baselines and our models with the following hyper-parameters. According to the scale of the dataset, the vocabulary sizes for OpenSubtitles, DailyDialog, PersonaChat, and MultiWOZ are set to  50k, 20k, 20k, and 18k, respectively. We use separate word embeddings for the encoder and the decoder, and the word embedding dimension is 256. All the parameters are initialized randomly from a normal distribution $\mathcal N (0, 0.0001)$. All models are trained using Adam \citep{DBLP:journals/corr/KingmaB14} with a learning rate of 0.001 and gradient clipping at 2.0. The batch size is 128. The hyper-parameters in our proposed appraoch are set as $\alpha = 0.01$ and $\lambda_{1} = 2.0$. Our models, i.e., \textbf{RegDG}, the imitating student conventional model, and \textbf{Transformer-IF}, the imitated teacher scenario-based model, are based on Transformer.

\begin{table*}[!t]
\centering
\scriptsize
\setlength{\tabcolsep}{7.0pt}

\begin{tabular}{@{}lccccccccccc@{}}
\hline 
\bf DailyDialog & Dist-1 $\uparrow$ & Dist-2 $\uparrow$ & Dist-3 $\uparrow$ & $D_{kl}^{u}$ $\downarrow$ & $D_{kl}^{b}$ $\downarrow$ & PPL $\downarrow$ & BLEU $\uparrow$ & GRE $\uparrow$ & AVE $\uparrow$ & EXT $\uparrow$ & COH $\uparrow$ \\ \hline 
Seq2Seq+Att & 0.42 & 1.66 & 3.83 & 13.865 & 29.096 & 116.72 & 16.7 & 0.4729 & 0.5308 & 0.3131 & 0.6096 \\
VHRED+BOW & 0.95 & 3.34 & 6.93 & 12.933 & 28.366 & 111.91 & 17.9 & 0.4801 & 0.5214 & 0.3008 & 0.5981 \\
NEXUS & 0.92 & 3.45 & 7.48 & 13.204 & 28.615 & 114.85 & 18.6 & \bf 0.4827 & 0.5415 & 0.3105 & 0.6254 \\
Transformer & 0.65 & 1.69 & 2.81 & 19.222 & 34.002 & 90.36 & 15.9 & 0.4605 & 0.5174 & 0.2961 & 0.6010 \\
ReCoSa & 0.66 & 2.18 & 3.97 & 16.471 & 32.429 & 92.76 & 18.3 & 0.4528 & 0.5205 & 0.2965 & 0.5465 \\
CHMAM & 0.82 & 2.43 & 4.40 & 15.288 & 30.835 & 97.42 & 17.7 & 0.4486 & 0.5112 & 0.2901 & 0.5920 \\ \hline
RegDG & \bf 1.15 & \bf 4.45 & \bf 9.22 & \bf 10.983 & \bf 26.846 & \bf 80.61 & \bf 19.1 & 0.4820 & \bf 0.5477 & \bf 0.3178 & \bf 0.6324 \\ \hline \hline 
\bf PersonaChat & Dist-1 $\uparrow$ & Dist-2 $\uparrow$ & Dist-3 $\uparrow$ & $D_{kl}^{u}$ $\downarrow$ & $D_{kl}^{b}$ $\downarrow$ & PPL $\downarrow$ & BLEU $\uparrow$ & GRE $\uparrow$ & AVE $\uparrow$ & EXT $\uparrow$ & COH $\uparrow$ \\ \hline 
Seq2Seq+Att & 0.11 & 0.41 & 0.94 & 14.488 & 27.562 & 136.83 & 19.2 & 0.4898 & 0.5099 & 0.2599 & 0.5998 \\
VHRED+BOW & 0.25 & 0.76 & 1.55 & 12.772 & \bf 26.616 & 137.08 & 20.6 & 0.4952 & 0.4827 & 0.2622 & 0.5743 \\
NEXUS & 0.26 & 0.89 & 2.02 & 11.325 & 27.134 & 145.32 & \bf 21.3 & 0.4940 & 0.4923 & 0.2654 & 0.6015 \\
Transformer & 0.28 & 0.64 & 1.01 & 25.076 & 36.985 & 90.52 & 16.0 & 0.4614 & 0.4976 & 0.2449 & 0.5824 \\
ReCoSa & 0.25 & 0.76 & 1.26 & 13.039 & 28.505 & 196.09 & 20.9 & \bf 0.4953 & 0.4519 & 0.2321 & 0.4906 \\
CHMAM & 0.42 & 1.27 & 2.22 & 11.558 & 27.009 & 159.59 & 20.5 & 0.4909 & 0.4453 & 0.2569 & 0.5358 \\ \hline
RegDG & \bf 1.11 & \bf 4.39 & \bf 9.43 & \bf 10.352 & 26.641 & \bf 83.36 & 21.1 & \bf 0.4953 & \bf 0.5257 & \bf 0.2665 & \bf 0.6163 \\ \hline \hline 
\bf OpenSubtitles & Dist-1 $\uparrow$ & Dist-2 $\uparrow$ & Dist-3 $\uparrow$ & $D_{kl}^{u}$ $\downarrow$ & $D_{kl}^{b}$ $\downarrow$ & PPL $\downarrow$ & BLEU $\uparrow$ & GRE $\uparrow$ & AVE $\uparrow$ & EXT $\uparrow$ & COH $\uparrow$ \\ \hline 
Seq2Seq+Att & 0.09 & 0.47 & 1.28 & 9.125 & 22.437 & 88.35 & 22.1 & 0.5194 & 0.5161 & 0.3156 & 0.6216 \\
VHRED+BOW & 0.10 & 0.50 & 1.42 & 8.990 & 21.903 & 113.82 & 22.5 & 0.5265 & 0.5148 & 0.3072 & 0.6149 \\
NEXUS & 0.09 & 0.53 & 1.60 & 9.106 & 21.986 & 97.31 & 22.8 & 0.5291 & 0.5212 & 0.3096 & 0.6237 \\
Transformer & 0.07 & 0.33 & 0.75 & 11.229 & 26.480 & 88.56 & 21.0 & 0.5295 & 0.4952 & 0.3033 & 0.5911 \\
ReCoSa & 0.07 & 0.38 & 0.91 & 10.188 & 25.144 & \bf 84.26 & 22.5 & \bf 0.5349 & 0.5029 & 0.3126 & 0.4746 \\
CHMAM & 0.09 & 0.41 & 0.98 & 10.129 & 24.885 & 89.33 & 22.3 & 0.4792 & 0.4512 & 0.2542 & 0.5612 \\ \hline
RegDG & \bf 0.12 & \bf 0.61 & \bf 1.61 & \bf 8.278 & \bf 21.709 & 85.68 & \bf 22.9 & 0.5300 & \bf 0.5282 & \bf 0.3175 & \bf 0.6345 \\ 
\hline 
\end{tabular}
\caption{The automatic evaluation results at the lowest point of the validation loss. The proposed approach achieves substantial improvements across all the dialogue datasets. ``$\uparrow$'' means higher is better. ``$\downarrow$'' means lower is better.}
\label{experiment result: automatic evaluation}
\end{table*}

\subsection{Evaluation Metrics}

We conducted both automatic and human evaluation to compare the performance of the models. 

\myparagraph{Automatic Evaluation} 
The evaluation of open-domain dialogue generation has no well-defined automatic metrics. Thus, we employ two kinds of automatic metrics to evaluate all models. The reference-based metrics, perplexity (\textbf{PPL}), \textbf{BLEU} (\%) \citep{DBLP:conf/acl/PapineniRWZ02}, and the embedding metrics (including embedding average (\textbf{AVE}), embedding greedy (\textbf{GRE}), embedding extrema (\textbf{EXT})) \citep{DBLP:conf/emnlp/LiuLSNCP16}, and coherence (\textbf{COH}) \citep{DBLP:conf/emnlp/XuDKR18}, are widely adopted to reflect the \textit{grammaticality and semantic relevance} of the responses \citep{DBLP:conf/aaai/SerbanSLCPCB17,DBLP:conf/acl/CsakyPR19}. The count-based metrics, distinct (\textbf{Dist-\{1,2,3\}} (\%)) \citep{DBLP:conf/naacl/LiGBGD16} and KL divergence \citep{DBLP:conf/acl/CsakyPR19}, are used to evaluate the \textit{lexical diversity} and the \textit{distribution distance} of the responses \citep{DBLP:conf/emnlp/XuRL018,DBLP:conf/nips/ZhangGGGLBD18}. We report the unigram and bigram version of KL divergence, i.e., $\vec{D_{kl}^u}$ and $\vec{D_{kl}^b}$.  Please refer to Appendix~\ref{ap:metrics} for the detailed settings of automatic metrics.

\myparagraph{Human Evaluation}
We conducted human evaluation to assess the quality of response. We randomly selected 200 test examples from each dataset and asked three annotators to judge which generated response in each pair (RegDG and baseline) is better (i.e., win, lose or tie) in terms of \textbf{Diversity} (how much the generated response contains meaningful information), \textbf{Relevance} (how likely the generated response is coherent to both dialogue history and future conversation), and \textbf{Fluency} (how likely the generated response is from human). 

\subsection{Experimental Results}

\myparagraph{Automatic Evaluation} 

The results obtained at the lowest point of the validation loss are shown in Table \ref{experiment result: automatic evaluation}. Our proposed model significantly outperforms all baselines on all datasets. The LSTM-based baselines obtain better performance than Transformer-based baselines in terms of diversity, distribution distance, and relevance, while they lose in grammaticality. It suggests that the LSTM-based model still has a certain advantage in the loose coupling dialogue task. Compared with CHMAM, ReCoSa achieves higher scores on BLEU and embedding metrics but weaker results on Dist-\{1,2,3\} and KL divergence, which means that only extracting scenario information from dialogue history cannot provide sufficient semantic constraints to improve model performance across all metrics. Although NEXUS and VHRED+BOW enrich the latent variable and bring more diversity and relevance, they show a distinct decline in PPL. It verifies that our method not only effectively uses the implicit conversation scenario to boost the performance but also indeed transfers this advantage to the inference phase. The improvements of our model on all datasets are significant with \text{$p \leq 0.01$} (t-test). The results of MultiWOZ, reported in Appendix~\ref{ap:multiwoz}, show similar improvements.

\begin{table*}[!t]
\centering
\scriptsize
\setlength{\tabcolsep}{3.0pt}

\begin{tabular}{c|l|ccc|c|ccc|c|ccc|c}
\hline
\multirow{2}{*}{Datasets} & \multicolumn{1}{c|}{\multirow{2}{*}{vs. Models}} & \multicolumn{3}{c|}{Diversity} & \multirow{2}{*}{$\kappa$} & \multicolumn{3}{c|}{Relevance} & \multirow{2}{*}{$\kappa$} & \multicolumn{3}{c|}{Fluency} & \multirow{2}{*}{$\kappa$} \\ \cline{3-5} \cline{7-9} \cline{11-13}
 & \multicolumn{1}{c|}{} & Win (\%) & Lose (\%) & Tie (\%) &  & Win (\%) & Lose (\%) & Tie (\%) &  & Win (\%) & Lose (\%) & Tie (\%) &  \\ \hline
\multirow{4}{*}{DailyDialog} 
 & VHRED+BOW & 53.25 & 19.00 & 27.75 & 0.484 & 52.00 & 22.25 & 25.75 & 0.442 & 39.25 & 19.50 & 41.25 & 0.532 \\
 & NEXUS & 50.50 & 18.00 & 31.50 & 0.500 & 47.75 & 23.75 & 28.50 & 0.556 & 38.00 & 18.75 & 43.25 & 0.523 \\
 & ReCoSa & 45.75 & 19.50 & 34.75 & 0.568 & 47.75 & 25.75 & 26.50 & 0.360 & 29.50 & 22.75 & 47.75 & 0.371 \\
 & CHMAM & 38.25 & 23.25 & 38.50 & 0.572 & 43.75 & 26.00 & 30.25 & 0.374 & 28.50 & 20.25 & 51.25 & 0.451 \\ \hline
\multirow{4}{*}{PersonaChat} 
 & VHRED+BOW & 46.00 & 30.50 & 23.50 & 0.433 & 43.25 & 39.75 & 17.00 & 0.440 & 34.50 & 26.50 & 39.00 & 0.442 \\
 & NEXUS & 41.25 & 23.00 & 35.75 & 0.677 & 40.75 & 34.25 & 25.00 & 0.557 & 30.00 & 23.50 & 46.50 & 0.652 \\
 & ReCoSa & 44.25 & 26.75 & 29.00 & 0.491 & 36.50 & 35.50 & 28.00 & 0.510 & 35.50 & 18.75 & 45.75 & 0.476 \\
 & CHMAM & 40.25 & 31.75 & 28.00 & 0.374 & 38.50 & 35.75 & 25.75 & 0.449 & 33.50 & 26.75 & 39.75 & 0.418 \\ \hline
\multirow{4}{*}{Opensubtitles} 
 & VHRED+BOW & 47.50 & 25.75 & 26.75 & 0.464 & 41.25 & 28.75 & 30.00 & 0.430 & 47.50 & 19.75 & 32.75 & 0.496 \\
 & NEXUS & 42.50 & 33.25 & 24.25 & 0.529 & 35.50 & 34.50 & 30.00 & 0.468 & 45.25 & 23.50 & 31.25 & 0.455 \\
 & ReCoSa & 31.00 & 21.25 & 47.75 & 0.445 & 40.00 & 13.00 & 47.00 & 0.372 & 30.50 & 11.25 & 58.25 & 0.324 \\
 & CHMAM & 29.25 & 28.25 & 42.50 & 0.448 & 35.75 & 27.75 & 36.50 & 0.469 & 28.00 & 18.00 & 54.00 & 0.457 \\ \hline
\end{tabular}
\caption{The human evaluation results. Our model has higher percentages of Win than the baselines.}
\label{experiment result: human evaluation}
\end{table*}

\myparagraph{Human Evaluation} 

The results are shown in Table \ref{experiment result: human evaluation}. We only report the results of VHRED+BOW, NEXUS, ReCoSa, and CHMAM, which are more related to our work. From the results, we can observe that our model performs better than baselines in all datasets. In particular, our model obtains the most significant win-lose difference on diversity, which demonstrates that the implicit conversation scenario induces the response containing more tokens that are meaningful. We calculate the Fleiss's kappa \citep{fleiss1971measuring} to measure the inter-annotator agreement, and the results are mainly distributed in $[0.4,0.6]$ (i.e., moderate agreement range) with the significance \text{$p \leq 0.01$}.

\subsection{Experimental Analysis}

In this section, we further quantitatively analyze the effectiveness of future conversations and explore why imitation learning works and how to enhance it. For limited space, we select a set of complementary metrics, Dist-\{1,2,3\}, PPL, and BLEU, to report the results. The rest of the results is in Appendix~\ref{ap:analysis}.

\begin{table}[t]
\centering
\footnotesize

\begin{tabular}{|@{\hspace{2pt}}p{0.98\linewidth}@{\hspace{2pt}}|}
\hline 
\textbf{Dialogue history:} Well, I am. How much will that cost? // The pass is free. // I don't have to pay for anything? \\ 
\hline 
\textbf{Seq2Seq+Att:} I don't like it. \\
\textbf{VHRED+BOW:} Yes, I'm sorry. \\
\textbf{NEXUS:} Yes, we have to pay the cash, please. \\
\textbf{Transformer:} I don't know what I want to do. \\
\textbf{ReCoSa:} I'm going to check out this magazine. \\
\textbf{CHMAM:} I want to get a deposit. \\
\textbf{RegDG:} You need to pay for the monthly sticker. \\ \hline
\textbf{Future Conversation:} How much is the monthly sticker? // It's \$ 24 for each month. // I'll take the student bus pass. \\ 

\hline \hline 

\textbf{Dialogue history:} We're considering of ordering 200 computers, but I'm wondering about the price you'll possibly offer. // Our price will be not less than \$ 5000. // Your price is higher than I expected. Could you give us a little discount? \\ 
\hline 
\textbf{Seq2Seq+Att:} Yes, I'm afraid I'm going to get it. \\
\textbf{VHRED+BOW:} Yes, we have a credit card. \\
\textbf{NEXUS:} Yes, I need to order the price, but I need to pay the goods in the price.\\
\textbf{Transformer:} I see. I'll take it. \\
\textbf{ReCoSa:} Well, we have to pay a discount for our products. \\
\textbf{CHMAM:} Yes, we'll take a 20\%. \\
\textbf{RegDG:} I'm afraid I can't. We don't have any reduction of quality. \\ \hline
\textbf{Future Conversation:} But the price is always negotiable and you should consider our quantity of order. // Well, what would you suggest? // Could you make it \$ 4500. \\  
\hline 
\end{tabular}
\caption{Examples of the generated responses. The responses generated by our model imply the implicit conversation scenario and contain meaningful information.}
\label{Result Analysis: Case Study}
\end{table}

\subsubsection{Case Study} 
Table \ref{Result Analysis: Case Study} presents some generated responses. The responses generated by baselines are usually dull and meaningless, while the responses generated by our model show diverse and coherent semantic information that indicates distinct relations with those topics in future conversation. The improvements of our model demonstrate the effectiveness of implicit conversation scenarios and our imitation learning framework. Due to limited space, we provide more examples in Appendix~\ref{ap:casestudy}. 

\subsubsection{Ablation Study} 
We evaluate the performance of our method without fine-grained prediction imitation (\textbf{FPI}) or intermediate representation imitation (\textbf{IRI}). The ablation study results, reported in Table \ref{Result Analysis: Ablation Study}, show that both types of imitation are beneficial for knowledge transfer. Without IRI, the model converges to a weaker performance than RegDG.

\begin{table}[t]
\centering
\scriptsize
\setlength{\tabcolsep}{4.5pt}

\begin{tabular}{@{}lccccc@{}}
\hline
\bf Models & Dist-1 $\uparrow$ & Dist-2 $\uparrow$ & Dist-3 $\uparrow$ & PPL $\downarrow$ & BLEU $\uparrow$ \\ \hline
RegDG & 1.1 & 4.4 & 9.2 & 80.61 & 19.1 \\
\; - FPI & 0.7 & 3.3 & 7.1 & 94.44 & 18.0 \\
\; - IRI & 0.9 & 4.1 & 8.7 & 87.93 & 18.4 \\ \hline
\end{tabular}
\caption{Results of the ablation study.}
\label{Result Analysis: Ablation Study}
\end{table}

\subsubsection{Effect of Future Conversation}

\myparagraph{Effect of the Informativeness of Future Conversation}
We first investigate the situation under which future conversations benefit the generation of high-quality responses. Intuitively, if the true responses or the future conversations are general safe responses, the future conversions contribute little useful information to current dialogue, thereby playing a limited role in the current response prediction. Thus, we classify examples into two sets, i.e., \textbf{Uninformative} and \textbf{Other}. Specifically, Uninformative includes the examples in which the \textit{\{dialogue history, response\}} or the \textit{\{response, future conversation\}} is a many-to-one pair. Generally, the second sequence in many-to-one pairs is dull and meaningless \citep{DBLP:conf/acl/CsakyPR19}. To determine the many-to-one pairs, we need to judge whether sentences are of the same meaning and we adopt three measures, that is, whether if the strings match (\textbf{Exact Match}), the words overlap more than 80\% (\textbf{Word Overlap}), or the sentences are in the same embedding cluster (\textbf{Sent. Cluster}). For the detailed settings of the above strategies, please refer to Appendix~\ref{ap:single-pass}. 

Table \ref{result analysis teacher: 3-1-3; 1-1-1; three strategies} shows the results of Transformer-IF on DailyDialog. We can see that the average of all metric improvements of Transformer-IF on the Uninformative set is lower than the Other set, which verifies the assumption that the informativeness of the future conversation supplementing the conversation scenario is crucial to the proposed approach.

\begin{table}[!t]
\centering
\scriptsize
\begin{tabular}{@{}llll@{}}
\hline 
\bf Sets & Exact Match & Word Overlap & Sent. Cluster \\ \hline 
Uninformative & $\times$1.035 & $\times$1.047 & $\times$1.059 \\ 
Other & \bf $\times$1.072$\ast\ast$ & \bf $\times$1.078$\ast\ast$ & \bf $\times$1.080$\ast$ \\
\hline 
\end{tabular}
\caption{The average of improvements across all metrics on the Uninformative set and the Other set. ``$\ast$'' and ``$\ast\ast$'' indicate $p\leq0.05$ and $p\leq0.01$, respectively. }
\label{result analysis teacher: 3-1-3; 1-1-1; three strategies}
\end{table}

\begin{table}[!t]
\centering
\scriptsize
\setlength{\tabcolsep}{4.5pt}

\begin{tabular}{@{}lccccc@{}}
\hline 
\bf 1-1-1 & Dist-1 $\uparrow$ & Dist-2 $\uparrow$ & Dist-3 $\uparrow$ & PPL $\downarrow$ & BLEU $\uparrow$ \\ \hline 
Transformer & 0.1 & 0.4 & 0.6 & 121.93 & 16.4 \\
Transformer-IF & \bf 0.2 & \bf 0.6 & \bf 1.2 & \bf 120.43 & \bf 17.7 \\
Improvement & +0.1 & +0.2 & +0.6 & +1.50 & +1.3 \\ \hline 
\bf 3-1-3 & Dist-1 $\uparrow$ & Dist-2 $\uparrow$ & Dist-3 $\uparrow$ & PPL $\downarrow$ & BLEU $\uparrow$ \\ \hline 
Transformer & 0.6 & 1.6 & 2.8 & 90.36 & 15.9 \\
Transformer-IF & \bf 0.8 & \bf 1.9 & \bf 3.6 & \bf 87.24 & \bf 20.2 \\
Improvement & +0.2 & +0.3 & +0.8 & +3.12 & +4.3 \\ 
\hline 
\end{tabular}
\caption{Results on DailyDialog (1-1-1) and (3-1-3).}
\label{result analysis teacher: 3-1-3; 1-1-1; information content; partial results}
\end{table}

\myparagraph{Effect of the Capacity of Future Conversation} 
In order to demonstrate the impact of the information content of the implicit conversation scenario on model performance, we conducted the training and testing of both Transformer and Transformer-IF on DailyDailog (1-1-1) and DailyDailog (3-1-3), respectively. ``3-1-3'' represents that both dialogue history and future conversation consist of three turns, and response only contains one turn. ``1-1-1'' represents that all sequences in the training examples consist of one turn. 

The results are shown in Table \ref{result analysis teacher: 3-1-3; 1-1-1; information content; partial results}. Compared with the results on DailyDialog (1-1-1), both models on DailyDialog (3-1-3) achieve overall improvements. The absolute improvements in multi-turn conversation are higher than those in single-turn conversation, which means that Transformer-IF performs better when the implicit conversation scenario contains rich semantic information. Because the automatic metrics may still improve after the lowest point of validation loss \citep{DBLP:conf/acl/CsakyPR19}, the results of both models after 50 epochs of training are reported in Appendix~\ref{ap:analysis}. It can be observed that Transformer-IF still substantially outperforms Transformer across all metrics under this setting. 

\subsubsection{Effect of Imitation Learning} 
\label{Effect of Imitation Learning}

\begin{figure}[t]
    \centering 
    \vspace{-1\baselineskip}
    \includegraphics[width=1.0\linewidth]{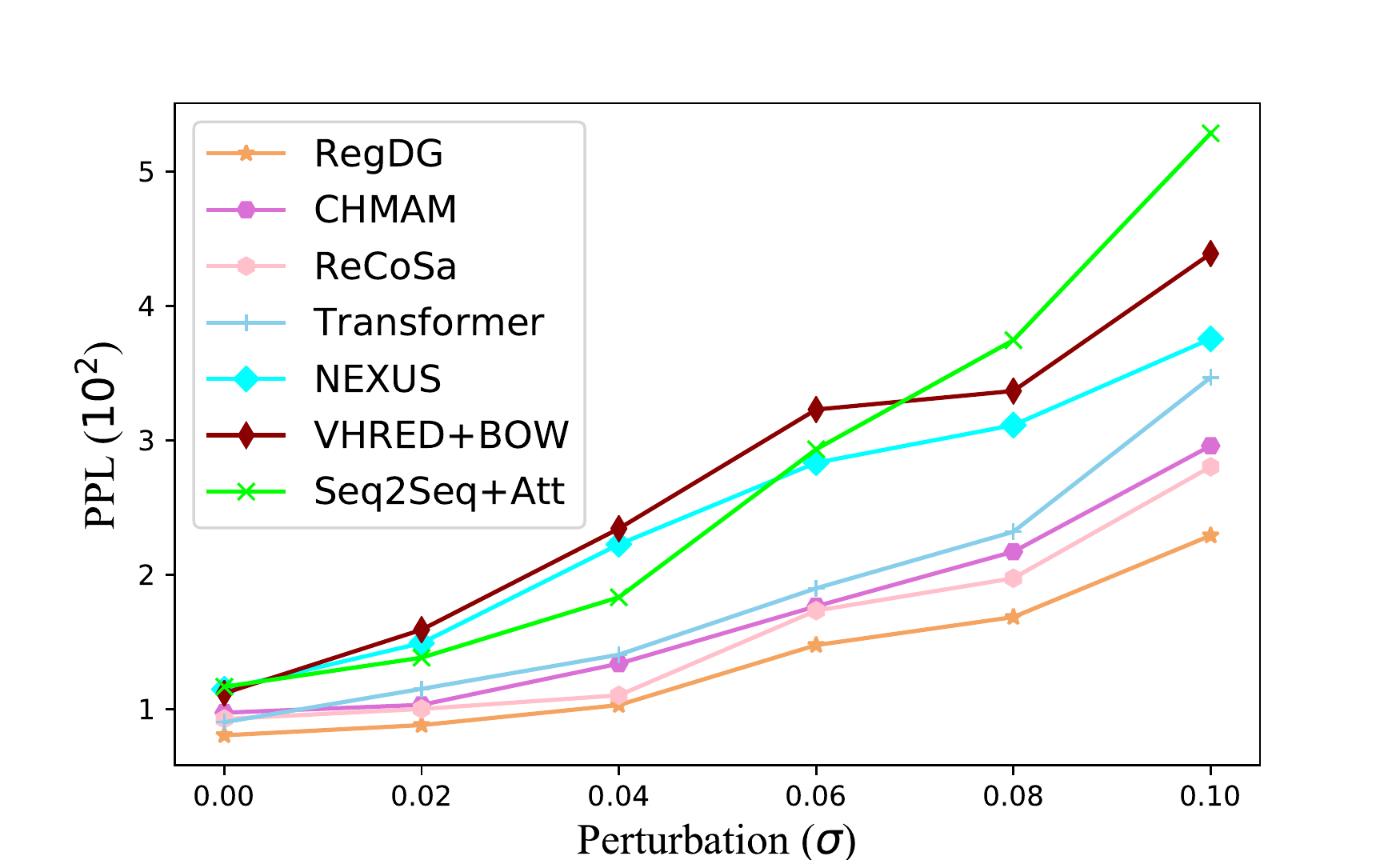}
    \caption{Analysis on model generalization.}
    \label{fig: Model Generalisation Analysis}
\end{figure}

\begin{table}[t]
\centering
\scriptsize
\begin{tabular}{@{}lc@{}}
\hline
\bf Models & Similarity \\\hline 
Seq2Seq+Att & 0.553 \\
VHRED+BOW & 0.689 \\
NEXUS & 0.700 \\
Transformer & 0.594 \\
ReCoSa &  0.633 \\
CHMAM &  0.656 \\
RegDG & \bf 0.732 \\ \hline 
\end{tabular}
\caption{Effect of regularization reflected by cosine similarity between the generated and real-world word distributions.}
\label{tab: Effect of regularization - word distribution}
\end{table}

\myparagraph{Why does the imitation learning work?} According to observations in previous work \citep{DBLP:conf/iclr/ChaudhariCSLBBC17,DBLP:conf/iclr/KeskarMNST17}, the model generalization is related to the width of the local minimum achieved by the model. Wider local minima suggests that the model can effectively resist perturbations and obtain better performance on unseen datasets. Therefore, we inject perturbations into the student to judge whether it is guided to a wider local minimum based on the regularization of knowledge transfer. Specifically, we add Gaussian noise with varying magnitude to the parameters of the trained model and observe the perplexity drop on the test set. The results in Figure~\ref{fig: Model Generalisation Analysis} show that the perplexity of all baselines rapidly increases while the perplexity of our student model grows slowly, indicating that the student model reaches a wider local minimum to gain better generalization. 

We also analyze the word distributions of the generated responses to intuitively reflect the effect of regularization from imitation learning. Concretely, we use a vector to represent all generated responses, and each element in the vector represents the frequency of a word. Only 2350 most frequent words are considered as \citet{DBLP:conf/aaai/FengCLY20}. Then, we calculate the distance between the word distributions from each model and the real-world data. From Table~\ref{tab: Effect of regularization - word distribution}, it can be seen that our model significantly outperforms plain Transformer and other baselines, which indicates that knowledge transfer effectively regularizes the model so that the model avoids sticking in a relatively centralized word distribution.

\begin{table}[!t]
\centering
\scriptsize
\setlength{\tabcolsep}{3.5pt}

\begin{tabular}{@{}lccccc@{}}
\hline 
\bf Models & Dist-1 $\uparrow$ & Dist-2 $\uparrow$ & Dist-3 $\uparrow$ & PPL $\downarrow$ & BLEU $\uparrow$ \\ \hline 
RegDG & 1.1 & 4.4 & 9.2 & \bf 80.61 & \bf 19.1 \\
\; + Word-Emb & 1.6 & 6.5 & 12.6 & 101.73 & 18.4 \\
\;\; + Encoder & \bf 1.8 & \bf 7.8 & \bf 17.7 & 118.00 & 17.9 \\
\hline 
\end{tabular}
\caption{Results using the hard transfer. With more hard-transferred modules, the diversity gradually improves, while the relevance gradually weakens.}
\label{Result Analysis IL: Can imitation learning be accelerated?; partial results}
\end{table}

\begin{figure}[!t]
    \centering 
    \includegraphics[width=1.0\linewidth]{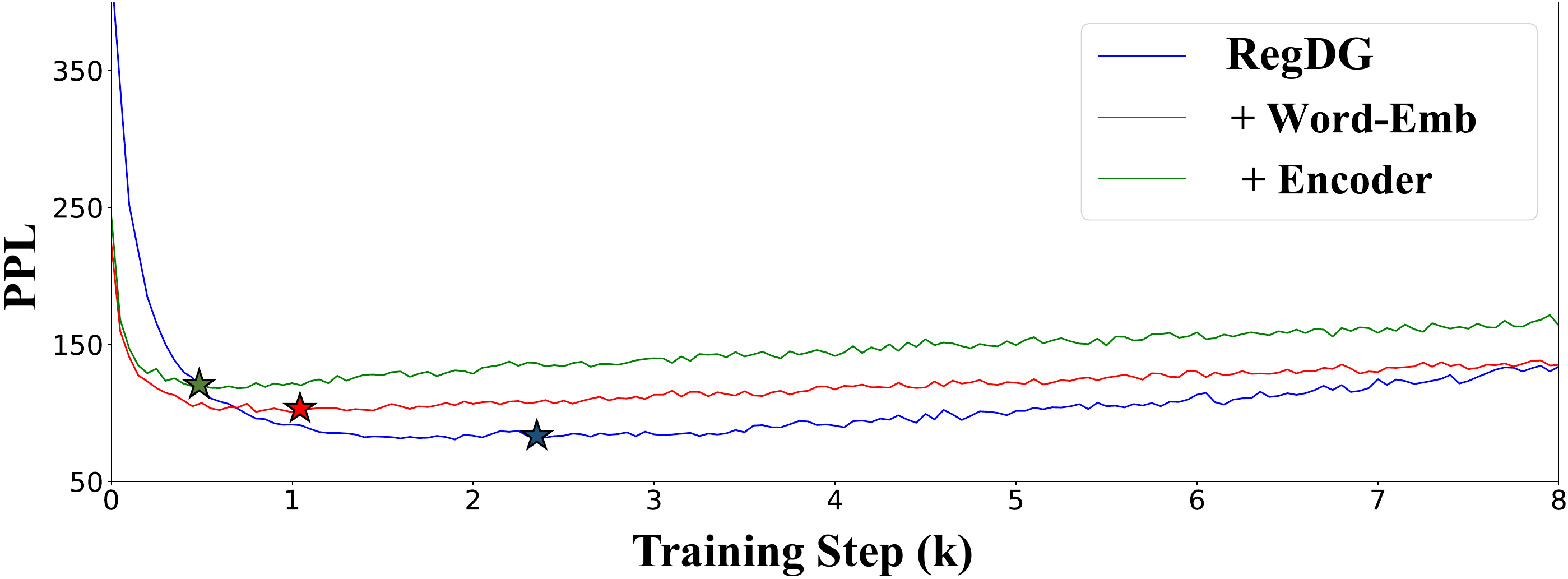}
    \caption{Convergence analysis of the hard transfer. The convergence is faster with more hard-transferred modules.}
    \label{fig: Convergence Analysis}
\end{figure}

\myparagraph{Can imitation learning be accelerated?} Before the student mimics the teacher, the teacher is usually well pre-trained. According to our observation, this is a redundant workflow that almost doubles the training time. It is worse if we should train a larger model on a huge dataset. In order to accelerate the training process, instead of transferring knowledge via supervised signals to train a student from scratch, we initialize the specified module of the student directly using the parameters of the teacher, and the transferred parameters are kept from updates during the training process. We call this the \textbf{hard transfer}. We first apply the hard transfer operation on word embedding (Word-Emb), and further extend it to the encoder. The results of DailyDialog in Table~\ref{Result Analysis IL: Can imitation learning be accelerated?; partial results} indicate that the performance has been further improved on the diversity with a slight drop on the relevance. Figure~\ref{fig: Convergence Analysis} shows the variation curve of PPL on the validation set with the training step. The full results are provided in Appendix~\ref{ap:analysis}. With more hard-transferred modules, the model reaches the lowest point of validation loss faster. It demonstrates that the hard transfer  distinctly accelerates the convergence.

\myparagraph{Do multiple teachers work?} To take advantage of more diverse and richer prior knowledge, we consider extending the teacher from one to many. We pre-train a transformer-based language model as another teacher. The results are shown in Table \ref{Result Analysis IL: Do multiple teachers work? partial results} with full results in Appendix~\ref{ap:analysis}. It is clear that with the help of the language model, the student further improves on all metrics, except for a weak decline in relevance, because the language model conducts unconditional sequence generation and does not consider the mapping between the dialogue history and the response. We defer the exploration of balancing multiple teachers in future work. 

\begin{table}[!t]
\centering
\scriptsize
\setlength{\tabcolsep}{4.5pt}

\begin{tabular}{@{}lccccc@{}}
\hline 
\bf Models & Dist-1 $\uparrow$ & Dist-2 $\uparrow$ & Dist-3 $\uparrow$ & PPL $\downarrow$ & BLEU $\uparrow$ \\ \hline 
RegDG & 1.1 & 4.4 & 9.2 & 80.61 & 19.1 \\
\quad + LM & \bf 1.3 & \bf 5.0 & \bf 9.6 & \bf 79.31 & \bf 19.4 \\ 
\hline 
\end{tabular}
\caption{Results of multiple teachers on DailyDialog. With the help of a pre-trained LM, the performance is improved consistently.}
\label{Result Analysis IL: Do multiple teachers work? partial results}
\end{table}

\section{Related Work} 

\myparagraph{Diversified Dialogue Generation} Recently, various researches have focused on neural dialogue models to generate diverse, informative, and relevant responses. 
One line of research attempts to extract relevant contexts from redundant dialogue history accurately \citep{DBLP:conf/aaai/XingWWHZ18,DBLP:conf/ijcai/TaoGSWZY18,DBLP:conf/acl/ZhangLPGC19}. 
Another line of research tries to explicitly incorporate a latent variable to inject the variability of response in the decoding process \citep{DBLP:conf/aaai/SerbanSLCPCB17,DBLP:conf/acl/ZhaoZE17}. \citet{DBLP:conf/emnlp/ShenSLK18,DBLP:conf/iclr/GuCHK19,DBLP:conf/naacl/GaoLZBGGD19} further enriched the latent variable approach. Also, some works redesigned the objective function or automatically learned it by adversarial learning \citep{DBLP:conf/naacl/LiGBGD16,DBLP:conf/emnlp/LiMSJRJ17,DBLP:conf/emnlp/XuRL018,DBLP:conf/aaai/FengCLY20}, which improves diversity but brings a fragile training process. Finally, some researchers have adapted external knowledge, such as topic information \citep{DBLP:conf/aaai/XingWWLHZM17}, persona \citep{DBLP:conf/acl/KielaWZDUS18}, knowledge base \citep{DBLP:conf/aaai/GhazvininejadBC18}. Unlike the above models to predict responses given a dialogue history, our method combines the future conversation with the dialogue history as the implicit conversation scenario, which contains comprehensive background information to guide the response generation.

\myparagraph{Imitation Learning} Imitation learning, acquiring skills from observing demonstrations, has proven to be promising in structured prediction, such as alleviating the exposure bias problem \citep{DBLP:conf/nips/BengioVJS15,DBLP:conf/acl/ZhangFMYL19}, transferring knowledge to guide non-autoregressive translation model \citep{DBLP:conf/iclr/Gu0XLS18,DBLP:conf/acl/WeiWZLS19}, and automatically learning the reward of the dialogue system \citep{DBLP:conf/aaai/LiKR19}. In our work, the conventional dialogue model as a student mimics the scenario-based dialogue model on both the output layer and intermediate layers.

\section{Conclusion} 
In this work, we introduce the future conversation with the corresponding dialogue history to learn the implicit conversation scenario, which entails latent context knowledge and specifies how people interact in the real world. To incorporate such scenario knowledge without requiring future conversation in inference, we propose an imitation learning framework. The scenario-based teacher model first learns to generate responses with access to both the future conversation and the dialogue history and then a conventional student model is trained to imitate the teacher by hierarchical supervisory signals. As a result, the student is effectively regularized to reach a robust local minimum that represents better generalization performance. Evaluation on four datasets demonstrates the effectiveness and the scalability of our approach, compared to the state-of-the-art baselines. The proposed framework enables the generation of responses that pertain more closely to the scenario indicated by the given dialogue history. Moreover, detailed analyses illustrate how imitating implicit scenarios regularizes the student model. For future work, we will incorporate pre-trained models into our framework (e.g., BERT as a teacher and GPT as a student) to further unlock the performance improvement and explore how to balance diverse prior knowledge from multiple teachers.

\section*{Acknowledgment}

This research is supported by Beijing Natural Science Foundation (No. L181010 and  4172054), National Key R\&D Program of China (No. 2016YFB0801100), and National Basic Research Program of China (No. 2013CB329605). This work is partly supported by Beijing Academy of Artificial Intelligence (BAAI). Xu Sun and Kan Li are the corresponding authors.

\bibliographystyle{acl_natbib}
\bibliography{emnlp2020}

\appendix

\section{Datasets}\label{ap:datasets}
\myparagraph{MultiWOZ} This dataset is a large-scale multi-turn conversation corpus that contains highly natural conversation across 7 goal-oriented scenarios written by human \citep{DBLP:conf/emnlp/BudzianowskiWTC18}. It is split to 58K, 15K, and 5K pairs for training, validation, and testing, respectively.

Two rules are applied to process the four datasets in the experiment: 
\begin{enumerate}
    \item Every seven consecutive dialogue turns form a training example, in which the first three turns, the middle turn, and the last three turns are taken as \textit{dialogue history}, \textit{response}, and \textit{future conversation}, respectively.
    \item The lengths of response, dialogue history, and future conversation are limited to $[5, 25]$, $[25, 80]$ and $[25, 80]$, respectively.
\end{enumerate}

\section{Baselines}\label{ap:baselines}
\paragraph{Seq2Seq+Att} We use a plain Seq2Seq model \citep{DBLP:conf/nips/SutskeverVL14} with attention mechanism \citep{DBLP:journals/corr/BahdanauCB14}. The encoder consists of a 2-layer bidirectional LSTM with 256 hidden units. The decoder is based on a 4-layer unidirectional LSTM with 256 hidden units. This baseline is enhanced with multiple techniques from related work and should be considered as a strong baseline.

\myparagraph{VHRED+BOW} VHRED is proposed by \citet{DBLP:conf/aaai/SerbanSLCPCB17}, which introduces conditional variational auto-encoder (CVAE) into the HRED model \citep{DBLP:conf/aaai/SerbanSBCP16} with a continuous latent variable attached to the response. We also adopt the BOW loss \citep{DBLP:conf/acl/ZhaoZE17} as a complementary with KL annealing. The latent variable is 256.

\myparagraph{NEXUS} NEXUS \citep{DBLP:conf/emnlp/ShenSLK18} enriches the latent variable with both dialogue history and future conversation through mutual information maximization.

\myparagraph{Transformer} Transformer \citep{DBLP:conf/nips/VaswaniSPUJGKP17} is based solely on the attention mechanism. The number of blocks and heads is 2 and 4, respectively. The hidden size is set to 256. The dimension of the feed-forward layer is 1024.

\myparagraph{ReCoSa} ReCoSa is proposed by \citet{DBLP:conf/acl/ZhangLPGC19}, which consists of a word-level LSTM encoder, a self-attention based context-level encoder, and a self-attention based context-response decoder. 

\myparagraph{CHMAM} CHMAM \citep{DBLP:conf/ijcai/TaoGSWZY18} applies Multi-Head Attention Mechanism (MHAM) to capture multiple semantic aspects from the dialogue history with a regularizer penalizing the redundancy of attention weight vectors across different aspects of the source sequence.

We adopt residual connection, Layer Normalization \citep{DBLP:journals/corr/BaKH16}, and Dropout in the LSTM-based baselines, which significantly boost the performance of Seq2Seq+Att, VHRED+BOW, and NEXUS. RegDG and Transformer-IF use the same settings as Transformer. The parameters of Transformer-IF are kept fixed during imitation learning.

\begin{table*}[!t]
\centering
\scriptsize
\setlength{\tabcolsep}{7.5pt}

\begin{tabular}{@{}lccccccccccc@{}}
\hline 
\bf MultiWOZ & Dist-1 $\uparrow$ & Dist-2 $\uparrow$ & Dist-3 $\uparrow$ & $D_{kl}^{u}$ $\downarrow$ & $D_{kl}^{b}$ $\downarrow$ & PPL $\downarrow$ & BLEU $\uparrow$ & GRE $\uparrow$ & AVE $\uparrow$ & EXT $\uparrow$ & COH $\uparrow$ \\ \hline 
Seq2Seq+Att & 0.18 & 0.73 & 1.62 & 4.269 & 14.332 & 20.82 & 23.2 & 0.5710 & 0.6557 & 0.4104 & 0.6520 \\
VHRED+BOW & 0.24 & 0.92 & 2.15 & 3.796 & 12.831 & 20.26 & \bf 24.4 & 0.5687 & 0.6555 & 0.3999 & 0.6546 \\
NEXUS & 0.28 & 1.01 & 2.16 & 4.020 & 13.656 & 18.64 & 23.1 & 0.5556 & 0.6339 & 0.4029 & 0.6547 \\
Transformer & 0.21 & 0.72 & 1.39 & 4.998 & 16.258 & 16.48 & 24.0 & 0.5634 & 0.6535 & 0.4085 & 0.6597 \\
ReCoSa & 0.20 & 0.79 & 1.61 & 4.567 & 14.684 & \bf 15.19 & 24.2 & 0.5731 & 0.6613 & 0.4044 & 0.5463 \\
CHMAM & 0.26 & 0.96 & 1.90 & 4.274 & 14.554 & 18.68 & 24.0 & 0.5612 & 0.6484 & 0.4151 & 0.6564 \\ \hline 
RegDG & \bf 0.27 & \bf 1.08 & \bf 2.37 & \bf 3.756 & \bf 12.744 & 16.23 & 24.2 & \bf 0.5738 & \bf 0.6619 & \bf 0.4234 & \bf 0.6601 \\ 
\hline 
\end{tabular}

\caption{The automatic evaluation results at the lowest point of the validation loss. ``$\uparrow$'' means higher is better. ``$\downarrow$'' means lower is better.}
\label{experiment result: automatic evaluation on MultiWOZ}
\end{table*}

\section{Automatic Evaluation Metrics}\label{ap:metrics}
\myparagraph{Embedding Metrics} The word embedding used in embedding metrics is pre-trained on each dataset using fasttext. 

\myparagraph{Perplexity} We calculate the perplexity of each individual sentence and then calculate the average perplexity, which is different from the other way that takes the whole test set (or each batch of data) as one long sentence and then calculates the perplexity (e.g., the official code of \citet{DBLP:conf/acl/ZhaoZE17}). The score calculated by the former way is usually larger than the latter one. The difference will increase as the test set (or the size of each batch) gets larger.

\myparagraph{KL Divergence} It calculates the KL divergence $D_{kl}=\nicefrac{1}{\left|U_{r}\right|} \sum_{w \in U_{r}} \log_{2} \nicefrac{p_{r}(w)}{p_{m}(w)}$ for each reference response between the reference and the generated response sets, where $p_{r}$ and $p_{m}$ are estimated based on the reference response set and the generated response set respectively.

\section{Automatic Evaluation on MultiWOZ}\label{ap:multiwoz}
Table \ref{experiment result: automatic evaluation on MultiWOZ} shows the automatic evaluation results on MultiWOZ. Our model beats all baselines on 9 out of 11 metrics.

\begin{table}[!b]
\centering
\scriptsize

\begin{tabular}{|@{\hspace{2pt}}p{0.98\linewidth}@{\hspace{2pt}}|}
\hline 
\textbf{Dialogue history:} Sure. The fitting room is on your right. // It fits well. I like it a lot. What do you think? // You look pretty in red. \\ 
\hline 
\textbf{Seq2Seq+Att:} I don't know. I don't want to do that. \\
\textbf{VHRED+BOW:} Oh, I don't know. \\
\textbf{NEXUS:} I don't know it. \\
\textbf{Transformer:} No, I don't think it's too heavy. \\
\textbf{ReCoSa:} I think I'll have a glass of wine. \\
\textbf{CHMAM:} It looks like this one. \\
\textbf{RegDG:} I'll take it. How much is it? \\ \hline
\textbf{Future Conversation:} \$ 29.99. // Okay, I'll take it. Thank you very much for your help. // You're welcome. \\ 

\hline \hline 

\textbf{Dialogue history:} First off, I'm going to need a pound of ground beef. // A pound of ground beef is \$ 2.48. // That's perfect. \\ 
\hline 
\textbf{Seq2Seq+Att:} How much do you need to do for the next week? \\
\textbf{VHRED+BOW:} How much do you want? \\
\textbf{NEXUS:} How much do you like? \\
\textbf{Transformer:} How much do you want to get? \\
\textbf{ReCoSa:} I'm going to need to get a ticket. \\
\textbf{CHMAM:} What kind of ground would you like? \\
\textbf{RegDG:} What kind of chicken do you want? \\ \hline
\textbf{Future Conversation:} I'm also going to need three pounds of chicken breasts. // The chicken breasts cost \$ 4.05 per pound. // What's the total price for the chicken? \\

\hline \hline 

\textbf{Dialogue history:} How much dose it cost for a bus pass? // It'll be \$ 65 for a monthly pass. // Is there anything cheaper than that? \\ 
\hline 
\textbf{Seq2Seq+Att:} Yes, I have to take a taxi. \\
\textbf{VHRED+BOW:} Well, it's the same thing. \\
\textbf{NEXUS:} That's right. How much do you want to pay the bus? \\
\textbf{Transformer:} I'm not sure. \\
\textbf{ReCoSa:} It's about two years ago. \\
\textbf{CHMAM:} We'll get a student pass. \\
\textbf{RegDG:} You can take a student pass. \\ \hline
\textbf{Future Conversation:} Well, I am. How much will that cost? // The pass is free. // I don't have to pay for anything? \\

\hline \hline 

\textbf{Dialogue history:} I don't want to have a specific title. I just want to enjoy what I am doing. // That sounds very reasonable. // It's the most important thing to me. \\ 
\hline 
\textbf{Seq2Seq+Att:} I don't know that. \\
\textbf{VHRED+BOW:} Well, I'll have a good idea. You can have a good idea. \\
\textbf{NEXUS:} What do you have to do? \\
\textbf{Transformer:} I don't know. But i don't know what to do. \\
\textbf{ReCoSa:} I think I should get a part-time job. \\
\textbf{CHMAM:} What kind of relationship do you have? \\
\textbf{RegDG:} But you know that that's a good idea. I think you should have a good impression on the job. \\ \hline
\textbf{Future Conversation:} That obviously depends on how things go. Whether I'm suited to the firm and firm to me. // Tell me about some of your recent goals and what you do to achieve them. // I want to put my knowledge and experience to use in a challenging position. In order to achieve this goal, I just want work step by step. \\  
\hline 
\end{tabular}

\caption{Examples of the generated responses.}
\label{Result Analysis: Case Study in the appendix}
\end{table}

\section{Case Study}\label{ap:casestudy}
We provide more generated examples in Table \ref{Result Analysis: Case Study in the appendix}.

\section{Experimental Analysis}\label{ap:analysis}

\paragraph{Effect of the Informativeness of Future Conversation\label{ap:single-pass}} We use the cosine similarity as the similarity measure and set 0.8 and 0.98 as the thresholds for one-turn sequence and three-turn sequence, respectively. The sentence embedding used in single-pass algorithm is the sum of word embedding with a corresponding weight estimated on the training set. We also adopt the k-means algorithm, which achieves similar clustering proportion.

\paragraph{Effect of the Capacity of Future Conversation} Table \ref{result analysis teacher: 3-1-3; 1-1-1; information content} demonstrates that our model works better when the implicit scenario contains rich information. Table \ref{result analysis teacher: 3-1-3; overfitting} indicates that our model still outperforms Transformer after a fixed epochs (50) of training. 

\paragraph{Effect of Imitation Learning} Table \ref{Result Analysis IL: Can imitation learning be accelerated?} and Table \ref{Result Analysis IL: Do multiple teachers work?} report the full version of the experimental results in Section \ref{Effect of Imitation Learning}.

\begin{table*}[!t]
\centering
\scriptsize
\setlength{\tabcolsep}{6.5pt}

\begin{tabular}{@{}lccccccccccc@{}}
\hline 
\bf 1-1-1 & Dist-1 $\uparrow$ & Dist-2 $\uparrow$ & Dist-3 $\uparrow$ & $D_{kl}^{u}$ $\downarrow$ & $D_{kl}^{b}$ $\downarrow$ & PPL $\downarrow$ & BLEU $\uparrow$ & GRE $\uparrow$ & AVE $\uparrow$ & EXT $\uparrow$ & COH $\uparrow$ \\  \hline 
Transformer & 0.1 & 0.4 & 0.6 & 20.559 & 34.814 & 121.93 & 16.4 & \bf 0.4688 & 0.4943 & 0.2889 & \bf 0.4943 \\
Transformer-IF & \bf 0.2 & \bf 0.6 & \bf 1.2 & \bf 17.664 & \bf 32.575 & \bf 120.43 & \bf 17.7 & 0.4636 & \bf 0.4978 & \bf 0.3000 & 0.4883 \\
Improvement & +0.1 & +0.2 & +0.6 & +2.895 & +2.239 & +1.50 & +1.3 & -0.0052 & +0.0035 & +0.0111 & -0.0060 \\ \hline 
\bf 3-1-3 & Dist-1 $\uparrow$ & Dist-2 $\uparrow$ & Dist-3 $\uparrow$ & $D_{kl}^{u}$ $\downarrow$ & $D_{kl}^{b}$ $\downarrow$ & PPL $\downarrow$ & BLEU $\uparrow$ & GRE $\uparrow$ & AVE $\uparrow$ & EXT $\uparrow$ & COH $\uparrow$ \\ \hline 
Transformer & 0.6 & 1.6 & 2.8 & 19.222 & 34.002 & 90.36 & 15.9 & 0.4605 & 0.5174 & 0.2961 & 0.6010 \\
Transformer-IF & \bf 0.8 & \bf 1.9 & \bf 3.6 & \bf 15.858 & \bf 31.455 & \bf 87.24 & \bf 20.2 & \bf 0.4658 & \bf 0.5388 & \bf 0.3063 & \bf 0.6227 \\
Improvement & +0.2 & +0.3 & +0.8 & +3.364 & +2.547 & +3.12 & +4.3 & +0.0053 & +0.0214 & +0.0102 & +0.0217 \\ 
\hline 
\end{tabular}

\caption{The results on DailyDialog (1-1-1) and DailyDialog (3-1-3).}
\label{result analysis teacher: 3-1-3; 1-1-1; information content}
\end{table*}

\begin{table*}[!t]
\centering
\scriptsize
\setlength{\tabcolsep}{6.5pt}

\begin{tabular}{@{}lccccccccccc@{}}
\hline 
\bf {3-1-3} & Dist-1 $\uparrow$ & Dist-2 $\uparrow$ & Dist-3 $\uparrow$ & $D_{kl}^{u}$ $\downarrow$ & $D_{kl}^{b}$ $\downarrow$ & PPL $\downarrow$ & BLEU $\uparrow$ & GRE $\uparrow$ & AVE $\uparrow$ & EXT $\uparrow$ & COH $\uparrow$ \\ \hline 
Transformer & 6.2 & 29.2 & 54.0 & 3.357 & 16.374 & 618.99 & 26.0 & 0.5225 & 0.5954 & 0.3716 & 0.6413 \\
Transformer-IF & \bf 6.3 & \bf 31.7 & \bf 60.7 & \bf 2.896 & \bf 14.831 & \bf 597.19 & \bf 31.3 & \bf 0.5539 & \bf 0.6320 & \bf 0.4027 & \bf 0.6574 \\ 
\hline 
\end{tabular}

\caption{The results on DailyDialog after 50 epochs of training.}
\label{result analysis teacher: 3-1-3; overfitting}
\end{table*}

\begin{table*}[!t]
\centering
\scriptsize
\setlength{\tabcolsep}{6.5pt}

\begin{tabular}{@{}lccccccccccc@{}}
\hline 
\bf Models & Dist-1 $\uparrow$ & Dist-2 $\uparrow$ & Dist-3 $\uparrow$ & $D_{kl}^{u}$ $\downarrow$ & $D_{kl}^{b}$ $\downarrow$ & PPL $\downarrow$ & BLEU $\uparrow$ & GRE $\uparrow$ & AVE $\uparrow$ & EXT $\uparrow$ & COH $\uparrow$ \\ \hline 
RegDG & 1.1 & 4.4 & 9.2 & 10.983 & 26.846 & \bf 80.61 & \bf 19.1 & \bf 0.482 & \bf 0.547 & \bf 0.317 & \bf 0.632 \\
\; + Word-Emb & 1.6 & 6.5 & 12.6 & 9.506 & 26.011 & 101.73 & 18.4 & 0.476 & 0.526 & 0.306 & 0.608 \\
\;\; + Encoder & \bf 1.8 & \bf 7.8 & \bf 17.7 & \bf 8.831 & \bf 24.776 & 118.00 & 17.9 & 0.475 & 0.541 & 0.315 & 0.623 \\ 
\hline 
\end{tabular}

\caption{The results about the hard transfer operation on DailyDialog.}
\label{Result Analysis IL: Can imitation learning be accelerated?}
\end{table*}

\begin{table*}[!t]
\centering
\scriptsize
\setlength{\tabcolsep}{6.5pt}

\begin{tabular}{@{}lccccccccccc@{}}
\hline 
\bf Models & Dist-1 $\uparrow$ & Dist-2 $\uparrow$ & Dist-3 $\uparrow$ & $D_{kl}^{u}$ $\downarrow$ & $D_{kl}^{b}$ $\downarrow$ & PPL $\downarrow$ & BLEU $\uparrow$ & GRE $\uparrow$ & AVE $\uparrow$ & EXT $\uparrow$ & COH $\uparrow$ \\ \hline 
RegDG & 1.1 & 4.4 & 9.2 & 10.983 & 26.846 & 80.61 & 19.1 & \bf 0.482 & \bf 0.547 & \bf 0.317 & \bf 0.632 \\
\quad + LM & \bf 1.3 & \bf 5.0 & \bf 9.6 & \bf 10.378 & \bf 26.369 & \bf 79.31 & \bf 19.4 & 0.475 & 0.535 & 0.311 & 0.616 \\ 
\hline 
\end{tabular}

\caption{The results about multiple teachers on DailyDialog.}
\label{Result Analysis IL: Do multiple teachers work?}
\end{table*}

\end{document}